\documentclass{article}

\usepackage{arxiv}

\usepackage{graphics}   
\usepackage{epsfig}     
\usepackage{amsmath}    
\usepackage{amssymb}    
\usepackage{hyperref}   
\usepackage{ifthen}     
\usepackage{subfig}

\newcommand{\propmethod}{{SCAIN}}
\newcommand{\interdependence}{{the interdependence between context and word interpretation}}
\newcommand{\Interdependence}{{The interdependence between context and word interpretation}}

\newcommand{\Section}{\section}
\newcommand{\Subsection}{\subsection}
\newcommand{\Subsubsection}{\subsubsection}

\title{SLAM-Inspired Simultaneous Contextualization and Interpreting for Incremental Conversation Sentences}

\author{
  Yusuke Takimoto \\
    Faculty of Science and Technology\\
    Keio University \\
    \texttt{takimoto@ailab.ics.keio.ac.jp} \\
\And
  Yosuke Fukuchi \\
    Faculty of Science and Technology\\
    Keio University \\
    \texttt{fukuchi@ailab.ics.keio.ac.jp} \\
\And
  Shoya Matsumori \\
    Faculty of Science and Technology\\
    Keio University \\
    \texttt{shoya@ailab.ics.keio.ac.jp} \\
\And
  Michita Imai \\
    Faculty of Science and Technology\\
    Keio University \\
    \texttt{michita@ailab.ics.keio.ac.jp} \\
}

\begin{document}
\maketitle

%
\begin{abstract}
Distributed representation of words has improved the performance for many natural language tasks such as machine translation and document classification.
In many methods, however, only one meaning is considered for one label of a word, and multiple meanings of polysemous words depending on the context are rarely handled.
Although research works have dealt with polysemous words, they determine the meanings of such words according to a batch of large documents.
Hence, there are two problems with applying these methods to sequential sentences, as in a conversation that contains ambiguous expressions.
The first problem is that the methods cannot sequentially deal with {\interdependence}, in which context is decided by word interpretations and the word interpretations are decided by the context.
Context estimation must thus be performed in parallel to pursue multiple interpretations.
The second problem is that the previous methods use large-scale sets of sentences for offline learning of new interpretations, and the steps of learning and inference are clearly separated.
Such methods using offline learning cannot obtain new interpretations during a conversation.
Hence, to dynamically estimate the conversation context and interpretations of polysemous words in sequential sentences, we propose a method of simultaneous contextualization and interpreting ({\propmethod}) based on the traditional simultaneous localization and mapping (SLAM) algorithm.
By using the {\propmethod} algorithm, we can sequentially optimize {\interdependence} while obtaining new interpretations online.
For experimental evaluation, we created two datasets: one from Wikipedia's disambiguation pages and the other from real conversations.
For both datasets, the results confirmed that {\propmethod} could effectively achieve sequential optimization of the interdependence and acquisition of new interpretations.
\end{abstract}



\Section{Introduction}
In natural language processing (NLP), distributed representation of words~\cite{hinton1984distributed} is a standard method that expresses word meanings in a vector space.
By using a distributed representation, it becomes possible to create models that extract relationships between words, and high performance with this approach has been achieved in many tasks such as machine translation~\cite{sutskever2014sequence, bahdanau2014neural, wu2016google} and document classification~\cite{socher2013recursive, kim2014convolutional}.
In a standard model using distributed representation, a single distributed representation is assigned for one label of a word.
It has been shown, however, that the accuracy of distributed representation is increased by extension over multiple meanings for one label, that is, by considering multiple distributed representations of polysemous words~\cite{li2015multi, trask2015sense2vec}.
This paper deals with interpretation problems for polysemous words with distributed representations.

Without extending the distributed representation to cover polysemous words, there is only one pair of correspondences between labels and the distributed representation vectors of words.
To determine the interpretation of a polysemous word, we need to consider context, including the content and flow of sentences, as well as the label information.
However, the context includes words occurring in the sentences.
To handle polysemous words, it is necessary to handle {\interdependence}, which means that context is determined by word interpretations, and the word interpretations are determined by the context.

There are several research works dealing with polysemous words in a distributed representation space.
These works proposed many ways to obtain multiple interpretation vectors: clustering distributed representation vectors into multiple interpretations~\cite{reisinger2010multi}, using semantic previous knowledge~\cite{iacobacci2015sensembed}, and counting the numbers of times of using interpretations and acquiring new interpretations while using major interpretations preferentially~\cite{li2015multi}.
Other works~\cite{liu2015topical, shi2017jointly} proposed methods of dealing with the interdependence between topics and polysemous word interpretations by using the topics of entire sentences.
All these methods achieve higher performance than that of models that are unaware of polysemous words, and they show the superiority of handling multiple distributed representations of polysemous words.

%
%
%
Even though the above research dealt with polysemous words in a distributed representation space, there are two serious problems of dealing with polysemous words in a sequential manner during a conversation.
First, all the previous methods use an entire document at one time to obtain a distributed representation for polysemous words, as opposed to sequential interpretation of polysemous words as sentences occur.
{\Interdependence} plays a significant role in dealing with polysemous words not only for document processing but also for conversation understanding.
Unlike documents, a conversation requires estimating the interpretations of polysemous words even with an incomplete context during the conversation, which is an issue in the process of constructing the context.
In addition, the context estimation must be performed in parallel with pursuing multiple interpretations because there is a possibility that a previously obtained interpretation is wrong and that another interpretation becomes a new candidate for the result as the conversation progresses.
The significant issue for a natural language algorithm to follow the content of a conversation is continuous updating of {\interdependence} for each sentence while keeping the candidate interpretations of polysemous words.
The second problem in dealing with polysemous words in a conversation is that the previous research uses an offline process in the stage of learning new interpretations.
This stage is clearly separated from the inference stage, and thus, it is not possible to obtain new interpretations online.
In contrast, a real human conversation operates differently when a person notices that she or he does not know a target object to which the conversation partner refers.
She or he can guess the meaning of the unknown interpretation from the content that has been spoken so far.
Unfortunately, previous methods that only obtain new interpretations offline cannot estimate new interpretations dynamically during a conversation.
%
%
%

%
In this paper, we propose an estimation method called simultaneous contextualization and interpreting ({\propmethod}), which estimates a conversation's context and interpretations of polysemous words dynamically in sequential sentences as in conversation. The {\propmethod} algorithm is based on the traditional algorithm of simultaneous localization and mapping (SLAM).
The SLAM algorithm is mainly used by mobile robots to probabilistically find their self-position and create an environment map while considering the relationship between the position and the map~\cite{thrun2005probabilistic}.
We adapt the SLAM algorithm because of the close similarity between the interdependence of the self-position and environment map in SLAM to that of the context and word interpretation in a conversation.
We specifically apply the algorithm of FastSLAM~\cite{montemerlo2002fastslam}, which is a kind of SLAM, to conversation situations in natural language.
{\propmethod} can sequentially optimize the {interdependence between context and word interpretation} and obtain new interpretations by generating vectors from the estimated context and word interpretations corrected by the context.
The contribution of this paper to the field of artificial intelligence is the introduction of the sequential optimization, which is essential for sharing recognition between people and machines in realtime interactions, and it leads to a general-purpose response system.

There are several steps, such as morphological analysis, syntactic analysis, meaning interpreting, dialog planning, and sentence generating in natural language processing.
In this paper, we tackle the meaning of interpreting part.
Therefore, we assume that the results of the morphological and syntactic analysis, which are generated by other software, can be used as the input for {\propmethod}.
We do not deal with dialog planning and generating sentences.

The structure of this paper is as follows.
Section 2 describes previous research, conversation situations that are difficult to handle with previous methods, and the traditional SLAM algorithm.
In section 3, we propose {\propmethod}, based on the SLAM algorithm.
In section 4, we experimentally evaluate the proposed method, and section 5 discusses the evaluation results.
Finally, in section 6, we describe our future works before concluding the paper in section 7.

\Section{Background}
\Subsection{Word Embedding}\label{ss_word_embedding}
Word2Vec~\cite{mikolov2013efficient} provides distributed representation of words~\cite{hinton1984distributed} with two-layer neural networks trained with a large-scale dataset, and it shows high performance of estimating word similarity.
Word2Vec involves two simple models: a skip-gram model and a continuous bag-of-words (CBOW) model. It acquires a high-dimensional vector representation of words that allows addition of the vectors without losing its linguistic meaning.
More efficient learning methods have also been proposed such as using topics of a whole sentence represented by a co-occurrence matrix of words~\cite{pennington2014glove}, using words' order of appearance~\cite{ling2015two}, and using the inflectional forms of words by handling subwords~\cite{bojanowski2016enriching}.

\Subsection{Polysemous Word Embedding}
Many pieces of researches on distributed representations have ignored polysemous words whose meanings vary depending on the context~\cite{mikolov2013efficient, pennington2014glove, ling2015two, bojanowski2016enriching}.
Some works, however, tried dealing with them in various ways~\cite{reisinger2010multi, huang2012improving, trask2015sense2vec, iacobacci2015sensembed, li2015multi, pelevina2017making, horn2017context, athiwaratkun2018probabilistic}.
One fundamental work achieved multiple vectors for one word by clustering words in a distributed representation space for each context and using the averages of the word vectors in the clusters as prototypes~\cite{reisinger2010multi}.
Another method used a weighted average of word vectors in sentences as an element representing a global context~\cite{huang2012improving}. It then learned distributed representations with neural networks by clustering the contexts and assuming that words in the different clusters have a different meaning.
Research on reducing the calculation cost of learning~\cite{trask2015sense2vec} achieved to obtain multiple and more effective representation vectors by using annotated labels instead of unsupervised clustering of contexts and reducing the numbers of learning and clustering word vectors.
A method of learning polysemous words using a semantic network~\cite{iacobacci2015sensembed} could learn multiple word vectors and calculate the similarity between words that have multiple interpretations by creating a corpus for each meaning of a word from BabelNet~\cite{navigli2012babelnet}.
A method using a Chinese restaurant process (CRP)~\cite{li2015multi} was proposed to achieve a way to obtain new interpretations of words sequentially.
This method accepts a new representation vector as a candidate interpretation if the probability of an interpretation based on the new vector becomes higher than that of the interpretation coming from the words surrounding the target word.
The probability of interpretation is multiplied by the number of times that the interpretation has been used so far after it becomes a candidate.
This frequency-based probability causes more frequently used interpretations to be used more.
The calculation of the probability of a new interpretation is different.
A constant $\gamma$ is multiplied by the probability of controlling the likelihood of obtaining a new interpretation.
In addition, a method to acquire multiple interpretations by unsupervised learning from existing word vectors~\cite{pelevina2017making} detects polysemous words by clustering graph nodes constructed from the semantic similarity between words.
Weighted averages of the word vectors in the clusters are used as new representation vectors.
A context encoder (ConEc)~\cite{horn2017context} is a method to express multiple interpretations of words in a single matrix by using negative samples to learn a Word2Vec model.
Multiplying the ConEc matrix by a context vector generates a representation vector matching the context.
Probabilistic FastText, which combines Gaussian Mixture with a distributed representation of words, expresses words as distributions instead of points and uses combinations of the distributions to support polysemous words~\cite{athiwaratkun2018probabilistic}.
ELMo~\cite{peters2018deep} appends context information to distributed representation of words.
The context is generated by Bidirectional LSTM, and their hidden weights are averaged to concatenate with existing word embeddings.
The extended embedding can represent multiple meanings of words.
The above methods comprise both learning and inference and require a sufficient amount of training data for learning.
Although the CRP-based method obtains new interpretations sequentially in an online manner, many of the other methods obtain them by clustering according to the context or by learning based on Word2Vec.
Moreover, they use offline processing, which requires all the training data.

\Subsection{Document Embedding}
Distributed representation has also been applied to documents.
Distributed representation of documents is a primary method to acquire a numerical representation of context, or what sentences mean.
The most fundamental method uses the average of word vectors in a document ignoring the appearance order and dependency of words.
Huan et al. proposed a method to use a weighted average of word vectors as a global context for neural networks, which provides more meaningful representations of sentences~\cite{huang2012improving}.
Doc2Vec~\cite{le2014distributed} learns a distributed representation of a document by giving ID information of the document to a Word2Vec model.
Doc2Vec was originally evaluated on the task of sentiment analysis.
It was later shown, however, that Doc2Vec has remarkable capabilities for other tasks~\cite{dai2015document} to express the contents of a document.
A weighted average using the inverse document frequency (idf) is based on weights corresponding to the frequency of word occurrence in a document~\cite{singh2015words}.
An idf-based weighted average can be used to efficiently acquire a distributed representation of a document with few dataset resources.
With extending the above ideas, a sparse composite document vector (SCDV)~\cite{mekala2016scdv} is obtained by taking representative points among word vectors corrected according to clusters of distributed representations and idf values.

Using a distributed representation to express an entire document makes it easier to calculate the similarity of documents and provides a basis for developing applications.
A simple average and an idf-based weighted average require small calculation cost to deal with sequential sentence processing. However, Doc2Vec expends high calculation cost because it needs to deal with an entire document for learning.
In addition, previous works depend on fixed word representations, which are acquired from pre-trained models.

\Subsection{Topic Estimation with Distributed Representation}
Research works on topic estimation have proposed useful ways to express the content of a document.
There are many techniques such as latent Dirichlet allocation (LDA)~\cite{blei2003latent}, latent semantic analysis (LSA)~\cite{landauer1997solution}, and non-negative matrix factorization (NMF)~\cite{lee2001algorithms}.
Among these methods, LDA is appropriate for estimating topics in human conversation because it provides the flexibility and advantages of topic coherence~\cite{stevens2012exploring}.

There is also research that extends LDA by using a distributed representation of words and improving the accuracy of topic estimation.
The use of a Gaussian distribution in the distributed representation space improves the categorical representation of LDA~\cite{das2015gaussian}, which enables dealing with words that do not appear in training sentences.
The original LDA assumes that a document is represented by a mixture of multiple topics and does not account for the relationships between words.
In contrast, the improved LDA using a distributed representation can account for these relationships.
Topical word embedding (TWE)~\cite{liu2015topical} can learn distributed representations of words and topics simultaneously.
It uses a method of predicting the embedding matrices of words and topics from around the target word by extending skip-grams.
It also obtains the representation vectors of words in consideration of topics through three learning methods: learning word and topic vectors independently, learning only word vectors, which also represent topics, and learning pairs of a word vector and a topic vector.
Based on the idea of TWE, Word-topic mixture (WTM)~\cite{xianghua2016improving} improves the accuracy of detecting distributed representations of words and topics by minimizing the topic distribution expected by LDA and the topic-word pair distribution obtained by TWE.
Latent topic embedding (LTE)~\cite{jiang2016latent} achieves mutual enhancement by integrating the learning of the distributed representation for words and topic estimation.
TopicVec~\cite{li2016generative} is a generative model that incorporates a distributed representation of words into LDA, and its performance exceeds that of existing methods for document classification tasks.
As another approach, skip-gram topical word embedding (STE)~\cite{shi2017jointly} tackles the problem of resolving the semantic ambiguity of a word by applying an expectation-maximization (EM) algorithm to the relation between word representation and topic estimation.
The idea of using an EM algorithm comes from the symmetry between the notions that a distributed representation of words is usually based on an estimated topic and that topic estimation is based on word representations.

Most of the methods use step-by-step iteration between estimating a topic and generating a distributed representation of words, and the mutual relationships above are not generally resolved simultaneously.
STE has an advantage, however, in optimizing the relationship at the same time by the EM algorithm.
All the above methods assume offline learning, and their processes require access to an entire document at one time.
Hence, we have to draw attention to the inappropriate characteristics of the previous research in dealing with conversations.

\Subsection{Limitations of Previous Research}
Here, we discuss two problems related to all research on the distributed representation of polysemous words, distributed representation of documents, and topic estimation dealing with polysemous words.
The first problem is that all previous methods require an entire document at one time to achieve a task, so they cannot deal with sequential input of sentences as in conversation.
In other words, no method can be applied to sequential input of conversational sentences while considering {\interdependence}.
The problem can be ignored for document processing because such models can use all the sentences of a document prior to execution, but this is impossible for online conversation processing.
The second problem is that acquiring new interpretations of polysemous words is also limited to an offline process in previous works.
It is thus currently impossible to acquire new interpretations online with sequential input of conversational sentences.

Except for the CRP-based method~\cite{li2015multi}, the target of all existing research on the distributed representation of polysemous words is a document.
In addition, many methods~\cite{trask2015sense2vec, reisinger2010multi, iacobacci2015sensembed, huang2012improving, pelevina2017making, horn2017context, athiwaratkun2018probabilistic} are techniques to acquire new interpretations offline, with the learning and inference stages clearly separated.
The separation of stages causes difficulty, in terms of computational complexity, with obtaining interpretations online through sequential sentences as in conversation.
On the other hand, the CRP-based method can handle sequential input of sentences.
It is difficult, however, for that method to correct a determined interpretation because it is determined according to the number of times it is used.

Furthermore, even though research on the distributed representation of documents provides ways to express long-term contexts~\cite{le2014distributed, dai2015document, mekala2016scdv}, it does not deal with the ambiguity of polysemous words, and the learning process is offline.
On the other hand, there is a method that can express sentences at a low cost by using weighted averages rather than through learning~\cite{singh2015words}.
That method depends, however, on an expression vector of words and does not consider {\interdependence}.
Research introducing distributed representations to topic estimation provides another way to express context.
Such research includes attempts at dealing with polysemous words by handling topics comprehensively~\cite{liu2015topical, shi2017jointly, xianghua2016improving}.
Except for STE~\cite{shi2017jointly}, however, these works also do not account for {\interdependence}.
Moreover, their learning processes are also offline.

\Subsection{Dynamic Estimaton Situations}
In this paper, we deal with the problem of dynamically estimating the context and interpretation of words in sequential input of sentences as in conversation.
There are two situations for such dynamic estimation, as we explain in the following subsections.
Both situations include everyday conversations, in which ambiguous interpretations frequently occur.

\Subsubsection{Parallel Context Estimation}
\label{sec_situation_para}
The first situation occurs when participants have to continue a conversation without resolving the interpretation ambiguity of polysemous words.
When a speaker's utterance contains an ambiguous expression such as a polysemous word, the listener disambiguates the expression by interpreting the word according to the conversation's previous content.
When the background knowledge of the listener differs from the speaker’s, however, she or he cannot uniquely determine the interpretation of an ambiguous expression at the initial stage of the conversation.
Instead, multiple interpretation patterns become possible for the listener.
As the conversation continues, the listener can obtain information affording more possibility to understand what the speaker wants to say.
The variety of interpretations of a polysemous word decreases with the progress of the conversation.
A temporarily determined interpretation may change to a more appropriate one as the conversation goes while the listener has not been able to obtain enough information from the speaker throughout the conversation.
In other words, an immediate determination is not an ideal way to determine the interpretation of an ambiguous word.
Rather, it is important to keep multiple candidate interpretations while estimating their appropriateness.

\Subsubsection{New Interpretation Development}
\label{sec_situation_new}
The second situation occurs when the listener assumes unknown knowledge as the interpretation of a word when noticing that she or he does not know what the word means.
As in the first situation where an ambiguous expression exists at the initial stage of a conversation, the listener assumes multiple interpretations, and the conversation progresses.
However, the expectations of all interpretations of the word can become low over the course of the conversation when the listener cannot accept any of the current candidate interpretations.
Then, she or he must develop a new interpretation imagined from the content of the speaker's utterances.

\Subsection{SLAM Algorithm}
SLAM is a method for navigation which constructs an environment map (usually two- or three-dimensional) and estimates a self-position simultaneously~\cite{thrun2005probabilistic}. The major applications can be seen in self-driving cars, uncrewed aerial vehicles, and planetary robots. SCAIN is namely inspired by the traditional SLAM algorithm to develop a novel approach to the conversation understanding settings.

In general, SLAM is solving interdependency which exists in map generation and self-localization from the observation; namely, to generate a map, a self-position is required, and vice versa.
When a mobile robot recognizes its surrounding environment, it uses devices such as a laser range finder, color camera, or depth camera. Since such sensors are attached to the robot, the measured results will be in the relative coordinates centered by the robot's positions. 
Generating a map from such observations requires the self-position since it involves the coordination from where the observation is obtained with egocentric-allocentric transformation.
The self-position can be instantly derived from the history of the robot's controls with the specification of components such as the diameter of the wheel and the location of the motors. However, when it moves in the real environment, the resulting position will be diverted from the actual position, due to the environmental influences such as wheel slip or collision. Therefore, localizing the self-position also requires the map to fix such an error as well.
Additionally, the SLAM considers two following errors caused by the nature of the real environment: the sensor error and the movement error. The sensor error occurs in the observation when sensing the environment, due to the physical characteristics of the sensing device or environmental noise. The movement error, which we mentioned before, is the error occurs due to the environmental influences.

\begin{figure}[h] \centering
\includegraphics[width=0.6\textwidth]{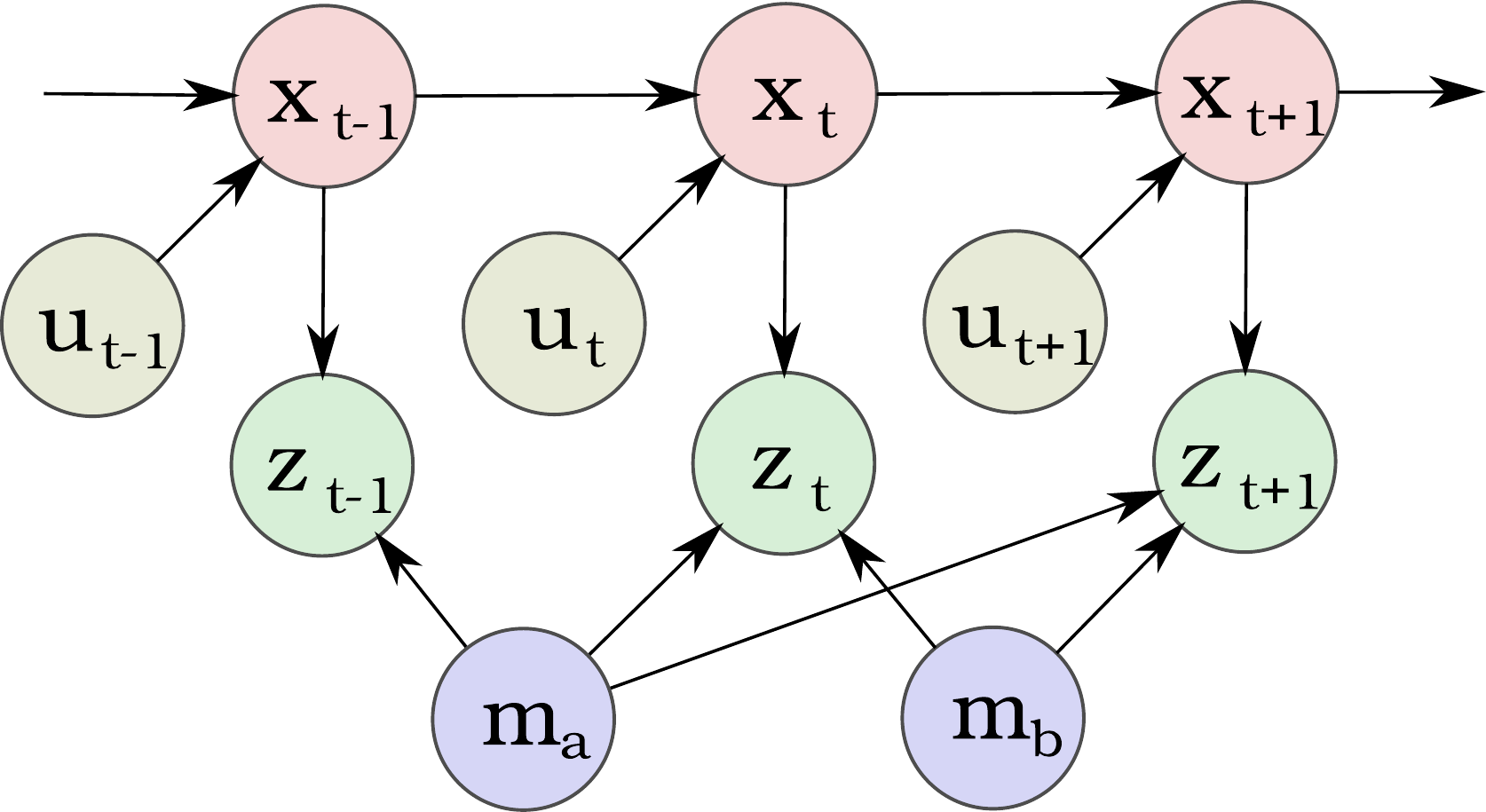}
\caption[Bayesian network representation of SLAM]{Bayesian network representation of SLAM, showing the interdependence among the self-position $x$, environment maps $m_a, m_b$, control $u$, and observation $z$.}
\label{fig_slam_network}
\end{figure}

The SLAM algorithm is formulized with four sets of random variables: the self-position $x$, environment map $m$, control $u$, and observation $z$.
The interdependency of these variables is expressed in the Bayesian network shown in Fig.~\ref{fig_slam_network}.
As the model shows, on each update at time $t$, the self-position $x_t$ is deriviated from the previous position $x_{t-1}$ according to the control $u_t$, and the observation $z_t$ is obtained from multiple environent maps $m_a, m_b, \ldots$ with respect to the self-position $x_t$.

\Subsection{FastSLAM}
We specifically use the FastSLAM algorithm~\cite{montemerlo2002fastslam} to develop our estimation method.
FastSLAM uses two types of filters: a particle filter and a Kalman filter.
It is categorized as both online SLAM and full SLAM, which means that it not only resolves the interdependence of localization and mapping sequentially depending on the robot movement but also fixes faults in the interdependence according to the full history of sensory and action data.
We chose FastSLAM because we believe that its characteristics are appropriate for handling the sequential input of a conversation and the dynamic update of word interpretation and context.
While there are several variations in FastSLAM, such as using landmarks or a grid map and knowing or not knowing landmarks' correspondences, we explain here the version of FastSLAM that uses landmarks whose correspondences are known because it is the most suitable variation for conversation situations.

\begin{figure}[h] \centering
\includegraphics[width=0.45\textwidth]{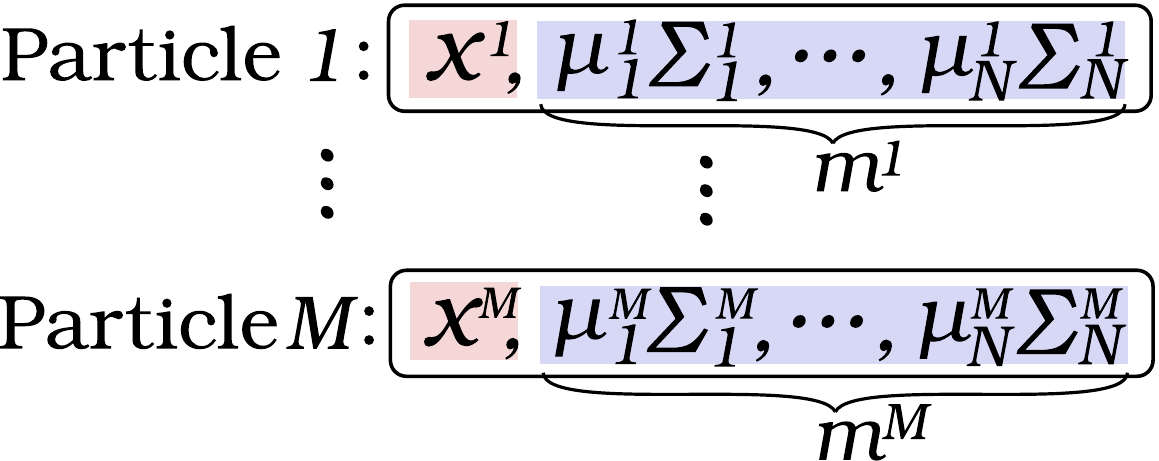}
\caption[Particle composition of FastSLAM]{Particle composition of FastSLAM: there are $M$ particles. Each particle has a self-position $x$ and an environment map $m$, which is represented by the existence probability distribution (average $\mu$, covariance $\Sigma$) of all $N$ observed landmark points.}\label{fig_slam_particle}
\end{figure}

FastSLAM consists of $M$ particles as shown in Fig.~\ref{fig_slam_particle}.
$M$ is decided by a tradeoff between the computational cost and the fineness of the search scale.
Each particle contains a self-position $x$ and landmark points which compose the environmental map $m$.
The landmark points are all the $N$ positions observed so far and are expressed as a Gaussian distribution with the average $\mu$ and covariance $\Sigma$ estimated by a Kalman filter.
Update processing is performed in the following three steps.

\Subsubsection*{Step1: Update Self-Position}
At the first step, the control $u$ is applied to the self-position $x$.
The self-position differs for each particle because it follows the existing probability distribution.
The self-position is updated by adding a noise according to the robot's motion model which estimates $p(x_t \textbar x_{t-1}, u_{t-1})$.
The specific distribution of the noise varies depending on the robot and the environment characteristics.

\Subsubsection*{Step2: Update Landmark Points}
In the second step, the observation $z$ is applied to the environment map $m$, and the landmark points are updated.
The version of FastSLAM considered here exploits the correspondences between landmark points observed at each time and those observed before that.
Updates are performed only when landmark points are observed.
Landmark points that were not observed continue to be the same as the previous ones.
The observed landmark points are updated by the Kalman filter and given the new average value $\mu$ and covariance $\Sigma$ of the new distribution.

\Subsubsection*{Step3: Resample Particles}
Finally, to incorporate the observation $z$ into the self-position $x$, resampling is applied via the particle filter.
The weight $w_t \propto p(z_t | x_t)$ is calculated for each particle, and each is regenerated with a probability proportional to the weight.
The probability distribution of $p(z_t|x_t)$ depends on the characteristics of the robot and environment, and it is usually designed according to the situation.
The existence of $M$ particles means that there are $M$ map patterns, and the particle with the largest weight $w_t$ at time $t$ is the best-estimated result at time $t$.

\Section{{\propmethod} Algorithm}
This paper proposes the {\propmethod} algorithm to dynamically estimate the context and interpretations of words in a conversation that includes polysemous words.
{\propmethod} leverages the FastSLAM algorithm to resolve {\interdependence}.
In particular, it achieves parallel context estimation while being able to obtain new information throughout the sequential input of a conversation text.

\Subsection{Association with SLAM}
The four random variables of the SLAM algorithm can be associated with a natural language conversation, as listed in Table~\ref{table_corresp}.
Through the association with natural language processing, the traditional SLAM algorithm offers a means of probabilistic processing for understanding context-dependent texts.
{\propmethod} has the same Bayesian network structure shown in Fig.~\ref{fig_slam_network} but with the random variables replaced according to the correspondences listed in Table~\ref{table_corresp}.

\begin{table}[htb]
\centering
\caption[Correspondence of random variables]{Correspondence of random variables between traditional SLAM and {\propmethod}.}
\begin{tabular}{c|c|c}
  Random variable & Traditional SLAM & Proposed {\propmethod} \\ \hline
  $x$ & Self-position & Context \\
  $m$ & Environment map & Interpretation domain \\
  $u$ & Control & Own utterance \\
  $z$ & Observation & Other's utterance \\
\end{tabular}
\label{table_corresp}
\end{table}

Whereas traditional SLAM focuses on a two- or three-dimensional space for the self-position $x$ and environment map $m$, {\propmethod} deals with a distributed representation space of words which we mentioned in subsection \ref{ss_word_embedding}.
The self-position $x$ of traditional SLAM is associated with the current context in a conversation.
In {\propmethod}, context $x$ is mapped on the distributed representation space as well as the words that appeared in the conversation.
This idea is similar to the one used in the global context~\cite{huang2012improving} and topic estimation~\cite{xianghua2016improving}.

The environment map $m$ corresponds to the interpretation domain indicating how to interpret words.
In an interpretation domain, each word in conversation is considered as a landmark, which is represented as a set of pairs of a label and its position.
Each pair expresses the relation between a word label and its interpretation in the distributed representation space.
A landmark position is expressed as a gaussian distribution with the average $\mu$ and covariance $\Sigma$, which represent the existence probability in the same manner as FastSLAM.

The control $u$ corresponds to the interpreter's own utterance, and the observation $z$ corresponds to the conversation partner's utterance.
{\propmethod} estimates the context $x$ based on the interpreter's own utterance $u$ and the other's utterance $z$.
The interpreter's own utterance $u$ has no ambiguity for themselves, but the conversation partner may wrongly interpret the utterance.
That is, the utterance can influence the context $x$ differently from what the interpreter intended.
Here, the movement error in SLAM corresponds to the uncertainty of the influence of the interpreter's own utterance $u$ on the context $x$ in {\propmethod}.

We consider the other's utterance $z$ as a result of observing words from the current context.
In SLAM, when a robot uses a laser range finder as a sensor, the observation $z$ is usually a distance measured by the laser beam.
The beam is emitted from the robot's position, which dynamically changes according to its control $u$ and reflected by an object in the environment.
By analogy with the reflected laser beam, we consider the other's utterance $z$ as the response to the interpreter's own utterance $u$ in terms of the domain $m$ and context $x$.
The other's utterance $z$ affects the context $x$ as well as $u$,
but $z$ is ambiguous for the interpreter in contrast to $u$.
Therefore, {\propmethod } treats $z$ as a probabilistic variable depending not only on the context $x$ but also on the interpretation domain $m$.

%

\Subsection{Procedure}

The procedure of {\propmethod} consists of four update steps.

\Subsubsection*{Step1: Update Context (1)}
In the first step, {\propmethod} applies the interpreter's own utterance $u$ to the context $x$.
Given the current utrance $u_t$ the update of the context $x_t$ is applied to each particle $k$
respectively via the following equation:
\begin{equation}
  \hat{x}^k_{t+1} = \left(1 - \lambda_u\right) x^k_{t} + \lambda_u \overline{v}_{u_t} + \sigma_u
  \label{eq_x_update_by_u}
\end{equation}
where $v_u$ is the average of the distributed representation of the words that compose the utterance $u$:

where $\lambda_u$ is a parameter representing the contribution of the utterances $u_t$ to the context $x$, $\sigma_u$ is a Gaussian random vector, and $\overline{v}_{u_t}$ is the average of the distributed representation of utterances $u_t$ deriviated as follows:
\begin{equation}
  \overline{v}_{u_t} = \frac{1}{N} \sum_{w \in u_t} v_w
\end{equation}
where $N$ is the number of words in the utterances $u_t$, and $v_w$ is the distributed representation of a word $w$ in $u$.
The distributed representation $v_w$ can be obtained from an arbitrary word embedding model such as Word2Vec~\cite{mikolov2013efficient} and GLOVE~\cite{pennington2014glove}.
In summary, the update operation (Eq.~\ref{eq_x_update_by_u}) moves the current context to the $\overline{v}_{u_t}$, with the error $\sigma_{u}$ which corresponds to the movement error in the SLAM algorithm.
Note that different particles have a different context; therefore, the updated contexts also differ within the particles.

\begin{figure}[t] \centering
  \subfloat[][Updating interpretation domain.]{\includegraphics[width=0.3\textwidth]{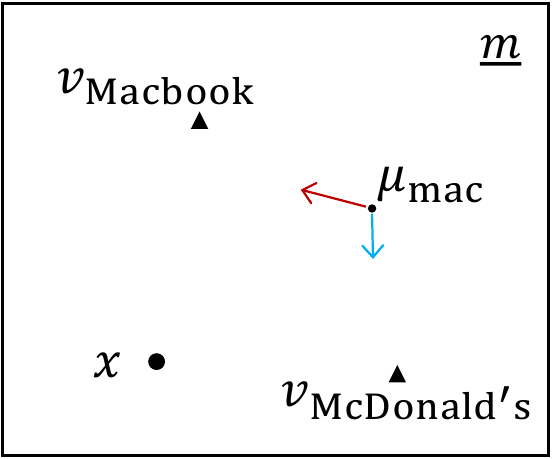}\label{fig_update_m}}
  \hspace{20pt}
  \subfloat[][Updating context.]{\includegraphics[width=0.3\textwidth]{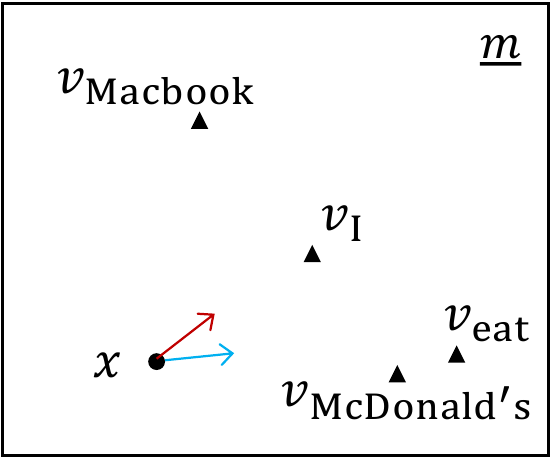}\label{fig_update_x}}
  \caption{The update examples of the SCAIN, given the conversation partner's utterance $z=$ ``I ate mac.’’. Since, the word `mac' is a polysemous word in Japanese, which can be either enterpreted as Macbook or McDonald's. Here, the interpretation of `mac' is conducted as follows: (Fig.~\ref{fig_update_m}) Interpreting each word in  $z$ based on the current interpretation ($\mu, \Sigma$), and current context $x$.
  }
\label{fig_scain_update}
\end{figure}

\Subsubsection*{Step2: Update Landmark Points}
Next, {\propmethod} interprets the other's utterance $z$ on the interpretation domain $m$ and updates the landmark points in the particles.
By updating the existence probability distribution of landmark points, interpretation domain $m$ is fixed according to the other's utterance $z$ so as to minimize the observation error contained in $z$.

{\propmethod} updates the average $\mu$ and covariance $\Sigma$ of each landmark point by applying the Kalman filter.
{\propmethod} considers that the landmark point of the word $w$ is observed at the position closer to the context $(1 - \lambda_w) x_t + \lambda_w v_w + \sigma_w$.
Here, $\lambda_w$ gives weight to the original position, and $\sigma_w$ is a Gaussian random number.
Adding noise toward the context corresponds to a sensor error in SLAM, which enables {\propmethod} to search more proper landmark position.
A landmark point is relocated toward the current context $x$ and its word vector $v$ (Fig. \ref{fig_update_m}).

Polysemous words, however, have multiple vectors for one label expression.
This ambiguity brings up the issue of finding an appropriate interpretation.
Imagine a sentence of the other's utterance $z$ that consists of $T$ words.
When there are $S_1, \ldots, S_T$ types of interpretations for each of the words, a sequence of pairs of a label $l$ and the distributed representation $v_l$, i.e., a sequence of candidates for interpretation, is represented by the following multiple patterns:
\begin{equation}
  \left<l_1 :
  \begin{cases}
    v_{l_1}^1 \\
    \vdots \\
    v_{l_1}^{S_1}
  \end{cases}\right>,
  \left<l_2 :
  \begin{cases}
    v_{l_2}^1 \\
    \vdots \\
    v_{l_2}^{S_2}
  \end{cases}\right>,
  \ldots,
  \left<l_T :
  \begin{cases}
    v_{l_T}^1 \\
    \vdots \\
    v_{l_T}^{S_T}
  \end{cases}\right>
\label{eq_label_vec_pattern}
\end{equation}
With respect to these multiple interpretation patterns, {\propmethod} updates the word vectors constituting the interpretation domain $m$ in each particle.
{\propmethod} applies these updates to all particles and all interpretation patterns.
A particle branches for each ambiguous word, and as a result, the number of particles increases per the number of interpretation candidates.

In FastSLAM with correspondences known, only observed landmarks are updated at each time, while non-observed landmarks are not updated.
{\propmethod}, however, has to deal with different update characteristics of conversational sentences in contrast to the update of measured distances for SLAM.
Specifically, the words that are obvious in a context are omitted.
In addition, words in an utterance can affect other words mentioned before and after it.
Therefore, we prepare a time window whose size is $t_\alpha$ to deal with these characteristics of conversational sentences.

\Subsubsection*{Step3: Update Context (2)}
In Step3, {\propmethod} reflects the effect of the other's utterance $z$ on the context $x$.
Fig. \ref{fig_update_x} illustrates the update process.
In normal SLAM settings, the position of a robot does not change without control.
However, the context in a conversation also changes by the other's utterance.
Therefore, {\propmethod} updates the context $x$ with the other's utterance $z$ in a similar way as Step1. Here, $\lambda_z$ is the contribution rate, and $\sigma_z$ is a Gaussian random number:
\begin{equation}
  x^k_{t+1} = \left(1 - \lambda_z\right) \hat{x}^k_{t+1} + \lambda_z \overline{v}_{z_t} + \sigma_z
  \label{eq_x_update_by_z}
\end{equation}

\Subsubsection*{Step4: Resample Particles}
As with the resampling step of FastSLAM, {\propmethod} calculates the weight $w_t \propto p (z_t | x_t)$ for each particle and resamples particles with probability proportional to the weight.
The weight $w$ of a certain particle at time $t$ is expressed via Eq. (\ref{eq_weight}) by using the context $x$ in each particle and $l_1: <\mu_{l_1}, \Sigma_{l_1}>, \ldots, l_N: <\mu_{l_N}, \Sigma_{l_N}>$, which is the existence probability distribution of the distributed representation updated from time $t-t_\alpha$ to $t$.
Here, $D_M$ is the Mahalanobis distance, representing the distance between the distribution and the vector, $\epsilon$ is a minimal value to avoid division by zero, and $\eta$ is an attenuation term to account for the time when the interpretation is updated (that is, $1$ means that the interpretation was used immediately before, while $0$ means that the interpretation has not been used for a long time):
\begin{eqnarray}
  w &=& -\log{\left( \frac{1}{N} \sum_{i=1}^{N}{\eta\left(l_i\right) D_M\left(<\mu_{l_i}, \Sigma_{l_i}>, x\right)} + \epsilon \right)}
  \label{eq_weight}
\end{eqnarray}

The weight $w$ is the negative logarithmic likelihood of the average of the distances, representing the degree of agreement between the context $x$ and the distribution of words in the recent utterances $z$ by the conversation partner.
{\propmethod} considers the characteristics of the occurrence of words to resample the particles.
A word that appears many times with another type of word has a large covariance because of the update of the Kalman filter.
For instance, words like ``thing'' and ``understand'' frequently appear within sentences.
As the large covariance makes it challenging to determine the position of the distributed representation, we use the Mahalanobis distance to account for the covariance in the weight calculation.
Typically, when calculating the similarity of words in a distributed representation space, cosine similarity is used to ignore the influence of the appearance frequency.
Because using the Mahalanobis distance in the Euclidean space in Eq. (\ref{eq_weight}) is affected by the appearance frequency, we convert the coordinate to an n-sphere coordinate~\cite{flanders1963differential} before calculating the distance.
The n-sphere is a multidimensional expansion of polar coordinates in two dimensions, and the distance in n-sphere coordinates corresponds to the angle in Euclidean coordinates.

We define the context $x_t$ and the interpretation domain $m_t$ of the particle with the largest weight $w_t$ as the most likely estimation.
The four-step update procedure of {\propmethod} makes it possible to obtain the best context $x$ in consideration of all the interpreter's own utterances $u_{1: t}$ and the other's utterances $z_{1: t}$ up to time $t$, as well as the interpretation domain $m_t$, which contains the distributed representation vectors whose meanings are corrected through conversation.

\Subsubsection*{New Interpretation Particle}
Although we can estimate the context and optimize existing interpretations with the above update steps of {\propmethod}, we cannot acquire completely new interpretations through these steps.
Thus, we introduce new interpretation particles to dynamically acquire new interpretations.

We assume that a word of unknown meaning in the other's utterance $z$ is not present in the interpretation domain $m$.
Hence, new interpretation particles are added to the particle set in Step2, and then weights are calculated and the particles are resampled together with existing particles in Step4.
The distributed representation of a word of unknown meaning is estimated from the past conversation content. The existence distribution of a new interpretation, $\mu_{new}, \Sigma_{new}$, is the average of recently used interpretations in the interpretation domain $m$ and context $x$.

New interpretation particles are resampled together with other particles in Step4.
To account for whether a new interpretation is necessary we replace the normal weight $w$ with the new weight $w'$ of the new interpretation particle.
Here, $w'$ is obtained by Eq. (\ref{eq_weight_new}):
\begin{eqnarray}
  w' &=& \gamma\left(t\right) \left( w + \lambda_{w2} \log{\left( \frac{1}{N_{l}} \sum_{j=1}^{N_{l}}{KL\left(<\mu_j, \Sigma_j> | <\mu_{new}, \Sigma_{new}>\right)} + \epsilon \right)} \right)
  \label{eq_weight_new}
\end{eqnarray}

The weight $w'$ of a new interpretation particle is modified from a normal weight $w$, specifically by adding a penalty term to subtract the degree to which the distribution of the new interpretation, $\mu_{new}, \Sigma_{new}$, matches the distribution of the existing interpretation vectors.
$w'$ becomes smaller as the new interpretation matches the existing ones, and it becomes larger if the new one has never existed before.
In the equation, $KL$ is the Kullback-Leibler (KL) divergence, representing the distance between two distributions. $N_{l}$ is the number of existing interpretation vectors for $l$, and $\lambda_{w2}$ is the contribution of the penalty term.
Lastly, $\gamma$ is a variable that controls the entire weight dependent on time $t$, which is influenced by the coefficient of the new interpretation probability obtained through a CRP process~\cite{li2015multi}. It increases as the conversation progresses, so that a new interpretation is less likely to occur at a stage with insufficient information at the beginning of the conversation.
Like the Mahalanobis distance, the KL divergence is calculated in n-sphere coordinates.

\Section{Experiments}
We conducted an experiment to evaluate the performance of the proposed {\propmethod} algorithm.
We focused on Japanese because it has many polysemous words.
Japanese mainly consists of 3 character types, Hiragana, Katakana, and Kanji.
Kanji characters contain complicated meanings, whereas Hiragana and Katakana express only pronunciation.
Since words written in Hiragana and Katakana always have the same spelling and pronunciation, there is no homonym that differs in pronunciation from the spelling which exists in European languages such as English.
Thus, because many homonyms can be treated as having multiple candidates of interpretation for a single label as in the case of polysemous words, the ambiguity of word interpretation is very high in Japanese, and it is suitable for the evaluation language.
However, {\propmethod} is not limited to Japanese, and in principle, it can be applied to other languages such as English.

As far as we know, there were no datasets that dealt with conversation sentences and topics temporally while focusing on ambiguous words.
Thus, we created two datasets for the evaluation.
The first was a Wikipedia dataset created from Wikipedia's disambiguation pages, and the second was a conversation dataset created from real human conversations.

The aim in a Wikipedia dataset was to evaluate whether {\propmethod} could interpret sequential sentences including ambiguous expressions and find proper referents.
Since we want to evaluate the {\propmethod}'s capability to resolve the ambiguous expression, we do not consider the effect of own utterances whose referents are clear. We treat the sentences in Wikipedia as if the interpreter's conversation partner is speaking for a long time.
In other words, the inputs towards {\propmethod} consist of only other's utterance $z$, and own utterance $u$ is not contained. Here, we emulate the situation where own utterance $u$ is just a simple response, and the conversation goes smoothly.
Other's utterance $z$ and own utterance $u$ update context $x$ based on similar equations (\ref{eq_x_update_by_u} and \ref{eq_x_update_by_z}).
In addition, the interpretation domain $m$ is updated by other's utterance $z$.
Therefore, we can evaluate the estimation of both context $x$ and interpretation domain $m$ by inputting other's utterance $z$.
We input sentences to {\propmethod} one by one and evaluated it at two different timing: at the time of inputting a sentence and after inputting all sentences.
We sought to confirm whether {\propmethod} could precisely find the referents indicated by the sentences at each of the two times.

We next aimed to evaluate whether {\propmethod} could correctly estimate the interpretations of polysemous words in an actual conversation between two humans and to examine the differences from human estimation results.
Specifically, we recorded how a participant interpreted ambiguous content spoken by the other participant in order to prepare sentences for input to {\propmethod}.
The second evaluation includes both of other's utterance $z$ and own utterance $u$, as opposed to the Wikipedia dataset.
In addition, because it contains numerical values of how topics are estimated during ambiguous conversations, it is possible to examine differences and similarities between human and {\propmethod}.

\Subsection{Experiment on Wikipedia Ambiguous Word Dataset}
\Subsubsection{Dataset Creation}
For sentences including ambiguous expressions, we used a Wikipedia dataset created from the disambiguation pages of Japanese Wikipedia.
Disambiguation pages are created to list referents for a word when there are multiple interpretations for the word expression.
The content of the page consists of sentences describing the meanings of the polysemous word and links to articles for the polysemous word in different meanings.
We selected the linked articles in terms of the different meanings of the polysemous word. We can know the correct meaning of the polysemous word in the article based on the link.

We selected 100 words at random from the contents of the disambiguation pages of Japanese Wikipedia. There was an average of about five kinds of interpretations per word, and the referent candidates of all the collected sentences totaled about 500 targets.
We arranged the order of the sentences at random to make it difficult for {\propmethod} to identify the referent of a polysemous word from sequential inputs of the sentences.
We made sure, however, that the first sequential sentence included a polysemous word in order to keep track of the estimated interpretations of the word at all times.
The sets of sentences were divided into two: one set for obtaining prior knowledge and adjusting the parameters for {\propmethod}, and the other for evaluation.
The number of sentences for evaluation was limited to thirty per article because it varied depending on the article.

\Subsubsection{Evaluation Method}
The evaluation was done with two tasks: an interpretation estimation task and a new interpretation task, respectively corresponding to the parallel context estimation explained in Section~\ref{sec_situation_para} and the new interpretation development explained in Section~\ref{sec_situation_new}.
In the interpretation estimation task, we input the sequential sentences describing a certain interpretation candidate for a polysemous word and evaluated whether the model could recognize it as a correct referent target.
In more concrete terms, we scanned all the particles generated by {\propmethod} and confirmed that it gave the word the same interpretation as in the meanings of the sentences. We also confirmed that the weight of the correct interpretation exceeded the weights of the majority of the other candidates, which meant that the weight ratio was sufficient.
In the new interpretation task, we evaluated whether {\propmethod} could obtain a new, unknown interpretation.
Similarly to the interpretation estimation task, sequential sentences were input, and the word interpretation was confirmed.
This task differed in that the correct interpretation was removed from the interpretation domain that {\propmethod} had.
The lack of the correct interpretation meant that {\propmethod} had to deal with sentences mentioning an unknown referent.
We thus confirmed whether the correct new interpretation could be added by new interpretation particles through the input of the sequential sentences.

As the initial value of the interpretation domain, we used a distributed representation obtained with FastText~\cite{bojanowski2016enriching}, which was learned from all the articles of Japanese Wikipedia except for the polysemous words' pages.
For the polysemous words, we created distributed representations from sentences that were not included in the dataset used for evaluation.
The hyperparameters for context update and weight calculation were optimized by a Gaussian process, and the parameters of the Kalman filter were optimized by the EM algorithm.
The number of particles was set to 20 times the number of interpretation candidates.
We used Juman++~\cite{tolmachev2018jumanpp} for morphological analysis of the Japanese sentences and KNP~\cite{kurohashi1994syntactic} for syntactic analysis, because the word breaks are not obvious and there is a need to be split it.

As a comparison condition to verify whether correction of the distributed representations was appropriate, we used a ``No Kalman'' model, which did not update the interpretation domain with the Kalman filter.  In addition, to verify whether the parallel search by the particle filter was necessary, we used a ``Fewer Particles'' model, in which the number of particles was reduced from 20 to 10 times the number of interpretation candidates.
These two comparison models were evaluated under the same conditions as {\propmethod} was.
Because the No Kalman model did not update the distributed representations from the initial values, it was equivalent to search by global context~\cite{huang2012improving}, which is simply weighted averaging.

\begin{figure}[htbp]
\begin{minipage}{1.0\hsize}
  \centering
  \includegraphics[width=1.0\textwidth]{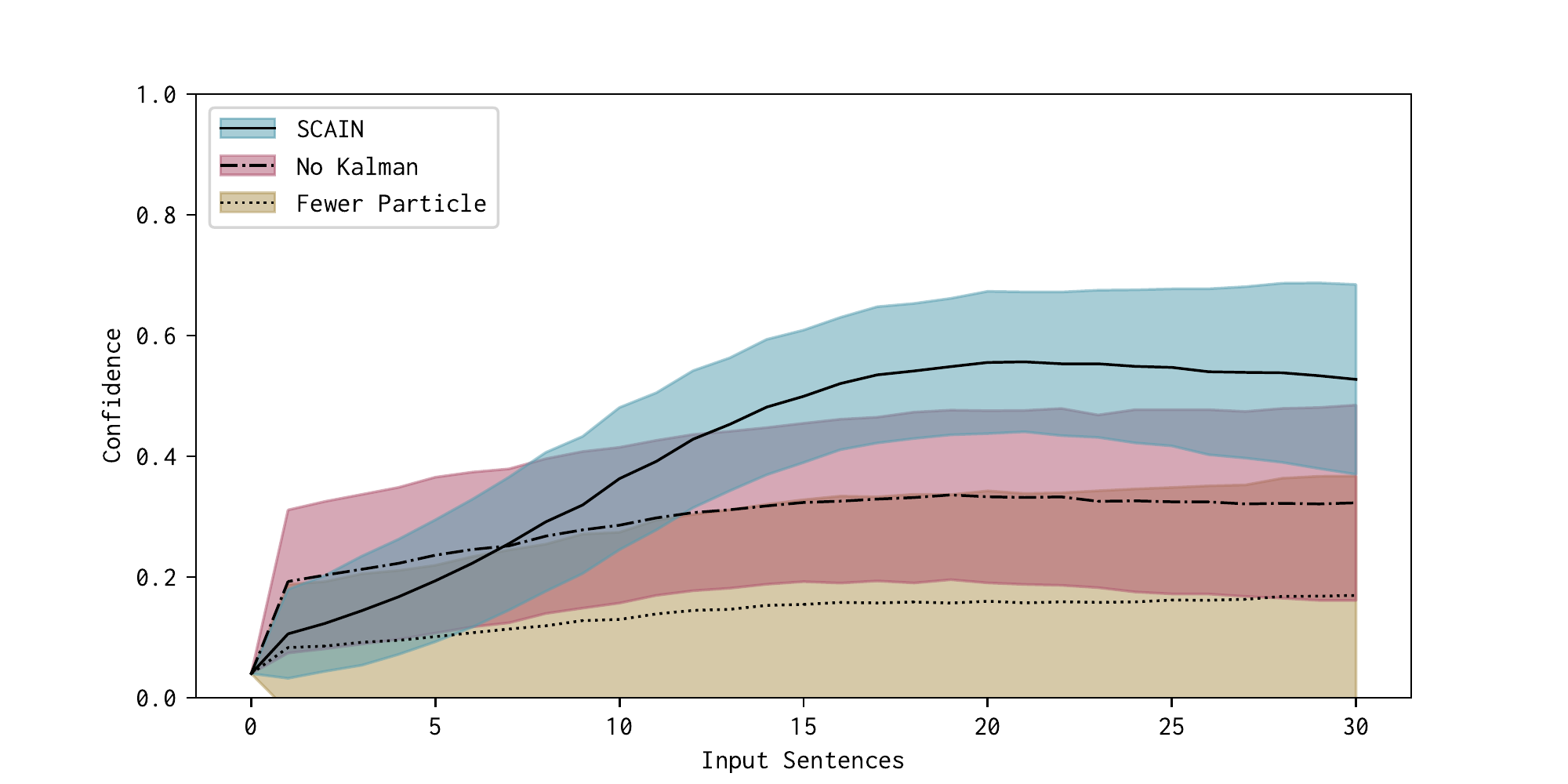}
  \caption[Expriment result of Wikipedia dataset's interpretaion estimation task]{Average and variance transitions of confidence with an increasing number of input sentences for the Wikipedia dataset on the interpretation estimation task.}
  \label{fig_eval_wiki_text}
\end{minipage}
\begin{minipage}{1.0\hsize}
  \centering
  \includegraphics[width=1.0\textwidth]{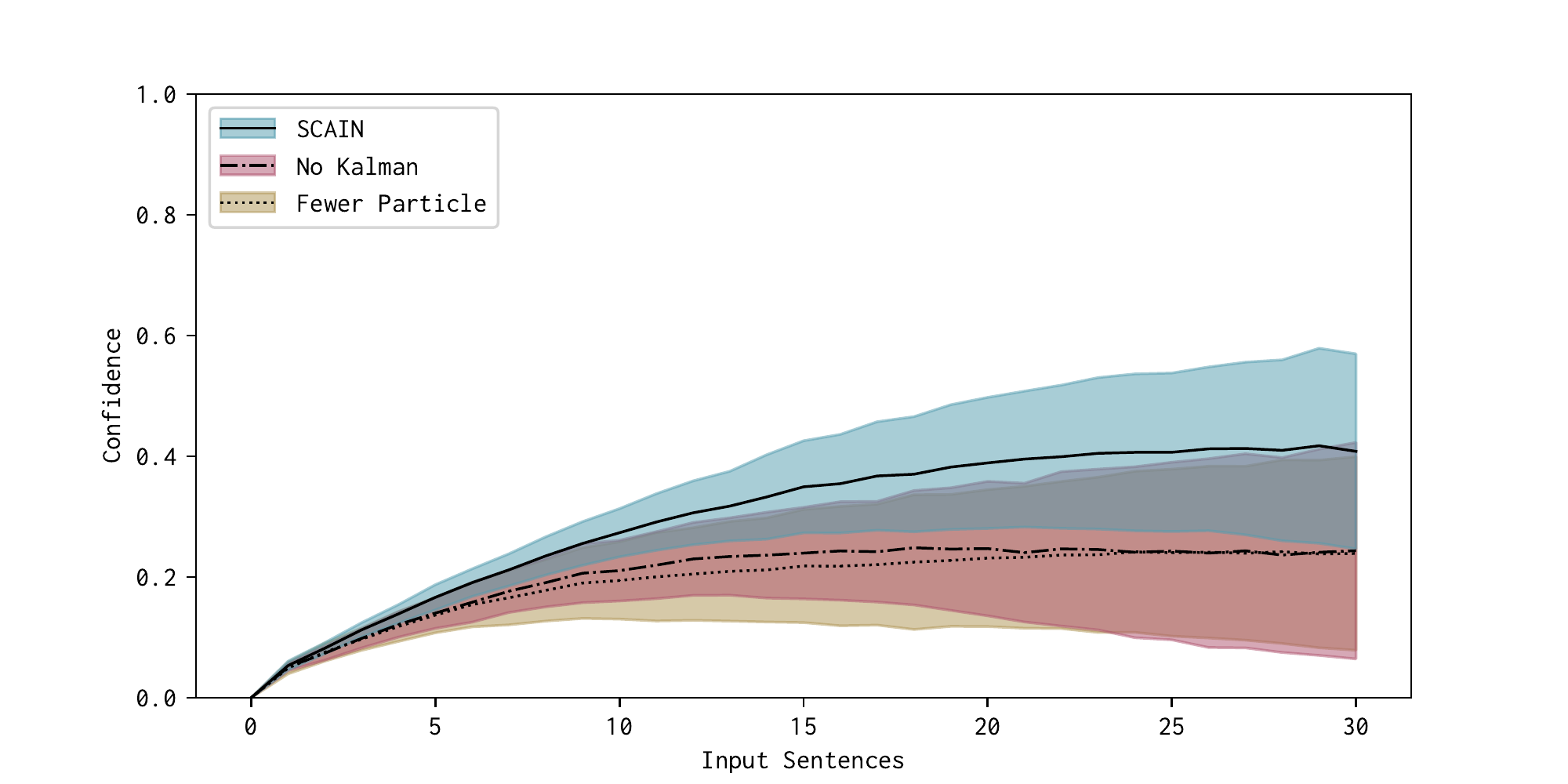}
  \caption[Expriment result of Wikipedia dataset's new interpretaion task]{Average and variance transitions of confidence with an increasing number of input sentences for the Wikipedia dataset on the new interpretation task.}
  \label{fig_eval_wiki_cand}
\end{minipage}
\end{figure}

\Subsubsection{Evaluation Result}
Figure~\ref{fig_eval_wiki_text} shows the transitions of the average and variance of the correct answer's confidence as the number of input sentences increased in the interpretation estimation task. Similarly, Fig.~\ref{fig_eval_wiki_cand} shows these transitions for the new interpretation task. Table~\ref{table_eval_wiki} lists the correct answer rates at the thirtieth input of the sentences for both tasks.
Here, the confidence is defined by Eq. (\ref{eq_confidence}) as the proportion represented by the largest weight among the group of particles classified by the estimated interpretations of a polysemous word.  Here, $\{w_{interp}\}$ is a group of weights whose particles give the same interpretation, and $\{w_{ans}\}$ is a group of weights with the correct interpretation:
\begin{equation}
  {confidence} = \frac{ \max(\{w_{ans}\}) }{ \sum_{interp}{\max(\{w_{interp}\})} }
  \label{eq_confidence}
\end{equation}

We can understand the confidence for a correct interpretation as follows.
The case of 0\% confidence means that all weights for the correct interpretation are zero, or that there is no particle estimating the correct interpretation.
In contrast, 100\% confidence indicates that all the weights of the other interpretations are zero while only the correct one has a weight, or that all particles estimate the correct answer.
The accuracy listed in Table~\ref{table_eval_wiki} is the proportion of correct interpretations whose final confidence after all sentences are input to {\propmethod} is more than 50\%, which means that the interpretation is the majority choice among the candidate interpretations.

All the confidence graphs in Fig.~\ref{fig_eval_wiki_text} started from the same value because all interpretation candidates had the same likelihood before inputting sentences.
The exact initial confidence was one divided by the number of interpretation candidates.
On the other hand, for the new interpretation task, Fig.~\ref{fig_eval_wiki_cand} indicates that the initial confidence of a new interpretation was zero, because there was no possibility that the new interpretation could become an interpretation candidate before inputting sentences to {\propmethod}.

\begin{table}[htb]
\centering
  \caption[Accuracy in the Wikipedia dataset]{Accuracy for the Wikipedia dataset, with the majority choice in terms of confidence taken as correct.}
\begin{tabular}{c|c|c}
  Model & Interpretation estimation task & New interpretation task \\ \hline
  {\propmethod} & 0.841 & 0.410 \\
  No Kalman     & 0.129 & 0.121 \\
  Fewer Particle & 0.166 & 0.127 \\
\end{tabular}
\label{table_eval_wiki}
\end{table}

\begin{figure}[h]
\centering
\includegraphics[width=0.4\textwidth]{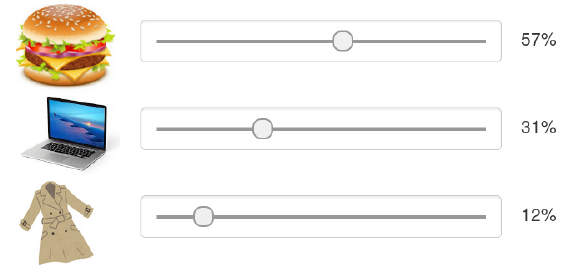}
\caption[Interface for inputting interpretation confidence]{Interface shown to estimators for inputting interpretation confidences in creating the conversation dataset. The example shows the polysemous word ``mac.''}
\label{fig_conv_slider}
\end{figure}

\Subsection{Experiment on Real Humans Conversation Dataset}
\Subsubsection{Dataset Creation}
We recruited external experimental collaborators and grouped them into pairs.
They talked to each other via a chat system on a computer so that we could collect conversation data.
The collaborators were 18 people, including 11 men and seven women, from 20 to 34 years old (average 22.78, variance 3.01). We paid 2,500 Japanese yen per person as an honorarium.
We asked one of the two collaborators to act as a presenter, who selects the object that is often referred to with the same polysemous word.
We then had the presenter talk about the objects for about five minutes per object.
We asked the other collaborator to be an estimator.
The estimator had to participate in the conversation while inputting the interpretation confidences of the ambiguous words spoken by the presenter.
For inputting the confidences, we showed multiple interpretation candidates on the computer together with sliders for each candidate, which represented a total value of 100\%.
Figure~\ref{fig_conv_slider} shows the input screen for the estimator.
In addition, because we wanted to make the conversation dataset always include the polysemous word throughout the conversation between the presenter and the estimator, we asked the presenter to say the polysemous word at the beginning of the conversation.
We also asked the presenter only to use the selected polysemous word to refer to the object, as we wanted to make it difficult for the estimator to determine the conversation's theme by using other words to express the object.
Even if the estimator was not sure what the presenter was talking about, we instructed them to continue the conversation so that the presenter would not notice the estimator's lack of comprehension.
As the theme objects consisted of ten kinds for one pair of experimental collaborators, the conversation dataset consisted of 90 conversations in total.
An example of a collected conversation is shown in Table~\ref{table_conv_dataset}, but translated into English.

\begin{table}[htb]
\centering
\caption[An example of conversation dataset]{Example of a conversation collected from the experimental collaborators. It consists of a conversation between a presenter (P) talking about the instructed theme and an estimator (E) estimating the theme. The example shows the case of interpreting the polysemous word ``mac'' as ``McDonald's.''}
\begin{tabular}{c|l|c}
  {} & \multicolumn{1}{c|}{Conversation} & Confidence of correct object \\ \hline
  P & I bought a mac. & --- \\
  E & Where did you buy it? & 33\% \\
  P & I bought it in Yokohama. & --- \\
  E & How was it? & 41\% \\
  P & There were a lot of people. & --- \\
  E & It seems popular. & 41\% \\
  P & Yes, it is popular. I ate three. & --- \\
  E & Do you go often? & 100\% \\
\end{tabular}
\label{table_conv_dataset}
\end{table}

\Subsubsection{Evaluation Method}
We evaluated {\propmethod} with the conversation dataset on the two tasks, the interpretation estimation task and the new interpretation task, as we had done with the Wikipedia dataset.
In both tasks, we input the sentences contained in the conversation dataset to {\propmethod} sequentially.
In the interpretation estimation task, {\propmethod} had all interpretation candidates as prior knowledge before the sentences were input, which is the same situation as the estimators knowing the candidates from the sliders on the computer.
On the other hand, in the new interpretation task, we did not give {\propmethod} an interpretation of the conversation theme to emulate the contribution of developing a new interpretation candidate.
{\propmethod} did have all the other candidates.
It is difficult, however, to obtain data from a legitimate situation in which the estimator does not have knowledge of a polysemous word that the presenter has, even though we wanted to trace the estimator's process of acquiring new interpretations of the polysemous word.
Thus, for the new interpretation task, we used the same values as in the interpretation estimation task by the human estimator as just a reference.

The interpretation candidates and the initial distributed representations for {\propmethod} were created from the Japanese Wikipedia articles as well as the evaluation with the Wikipedia dataset.
We limited the interpretation candidates to be the same as those shown to the estimator when creating the conversation dataset.
The Wikipedia articles were not divided into evaluation and learning sets, unlike the evaluation with the Wikipedia dataset, and all of them were used for learning.
In addition, the ``No Kalman'' and ``Fewer Particles'' models were used as comparison conditions, as in the case of the Wikipedia dataset.

\begin{figure}[htbp]
\begin{minipage}{1.0\hsize}
  \centering
  \includegraphics[width=1.0\textwidth]{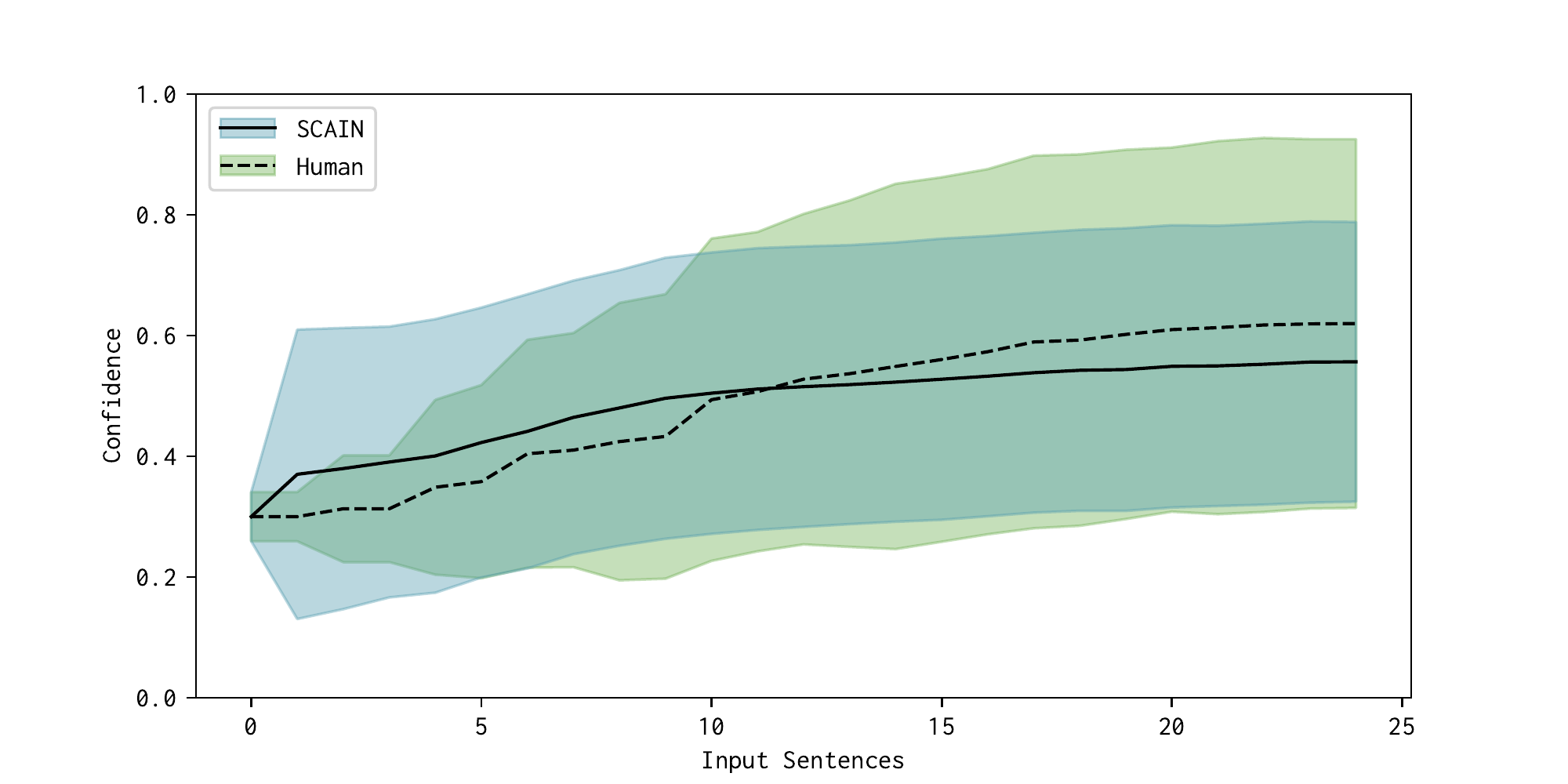}
  \caption[Expriment result of conversation dataset's interpretaion estimation task with human]{Average and variance transitions of confidence with an increasing number of input sentences for the conversation dataset on the interpretation estimation task, including comparison with human estimation.}
  \label{fig_eval_conv_text_human}
\end{minipage}
\begin{minipage}{1.0\hsize}
  \centering
  \includegraphics[width=1.0\textwidth]{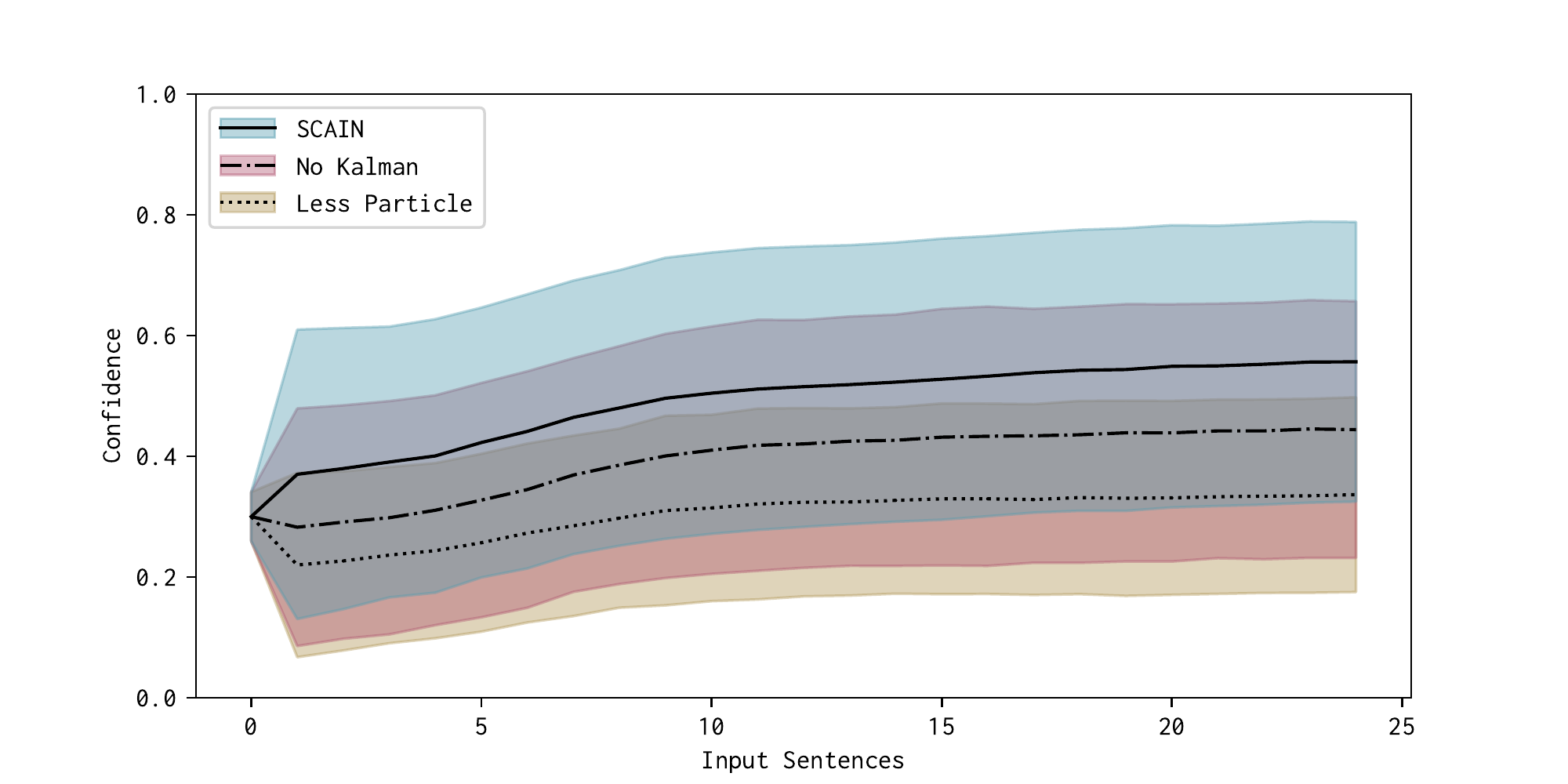}
  \caption[Expriment result of conversation dataset's interpretaion estimation task with other moels]{Average and variance transitions of confidence with an increasing number of input sentences for the conversation dataset on the interpretation estimation task, including comparison with the other models.}
  \label{fig_eval_conv_text_inner}
\end{minipage}
\end{figure}

\begin{figure}[htbp]
\begin{minipage}{1.0\hsize}
  \centering
  \includegraphics[width=1.0\textwidth]{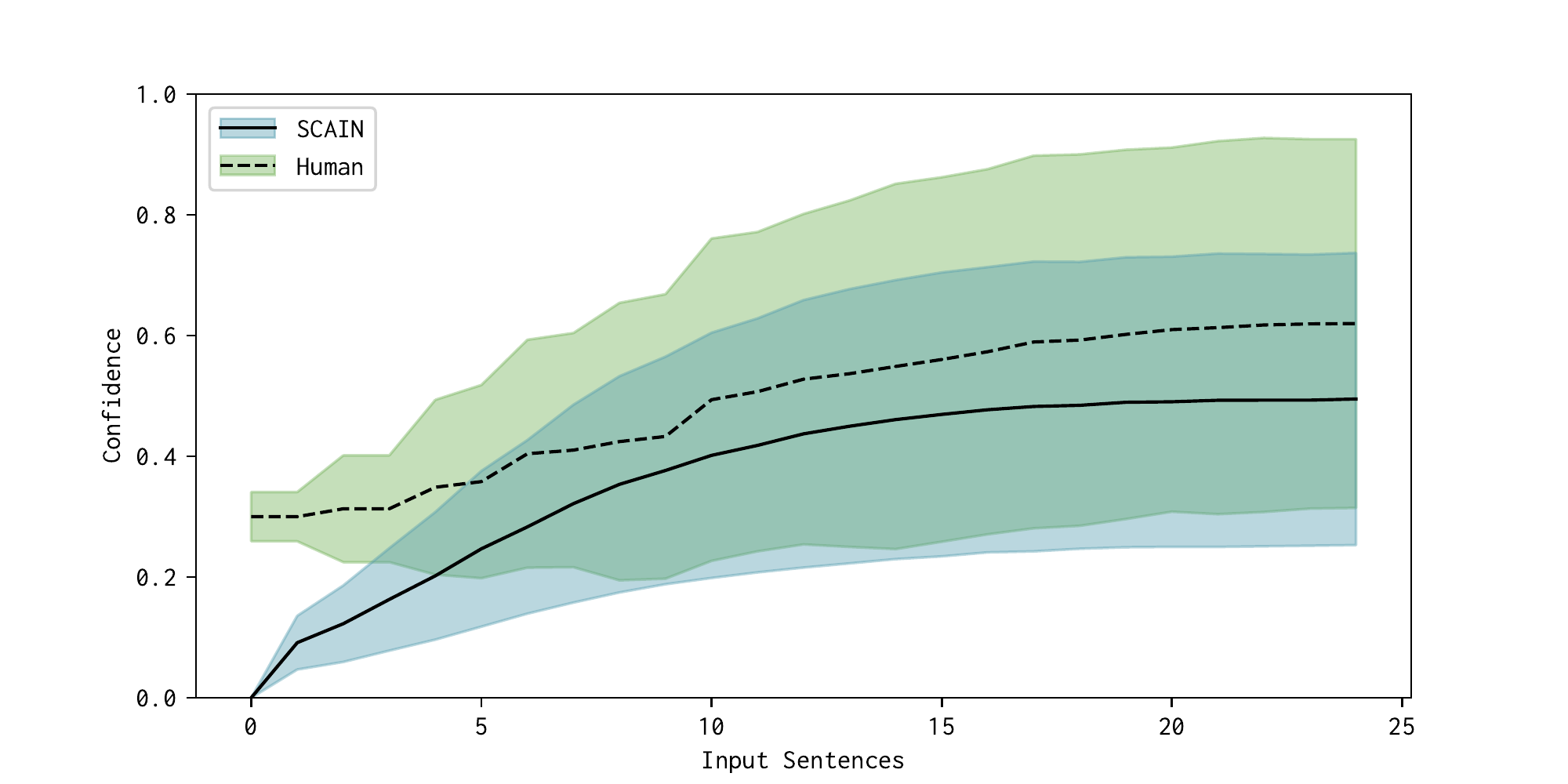}
  \caption[Expriment result of conversation dataset's new interpretaion task with human]{Average and variance transitions of confidence with an increasing number of input sentences for the conversation dataset on the new interpretation task, including comparison with human estimation.}
  \label{fig_eval_conv_cand_human}
\end{minipage}
\begin{minipage}{1.0\hsize}
  \centering
  \includegraphics[width=1.0\textwidth]{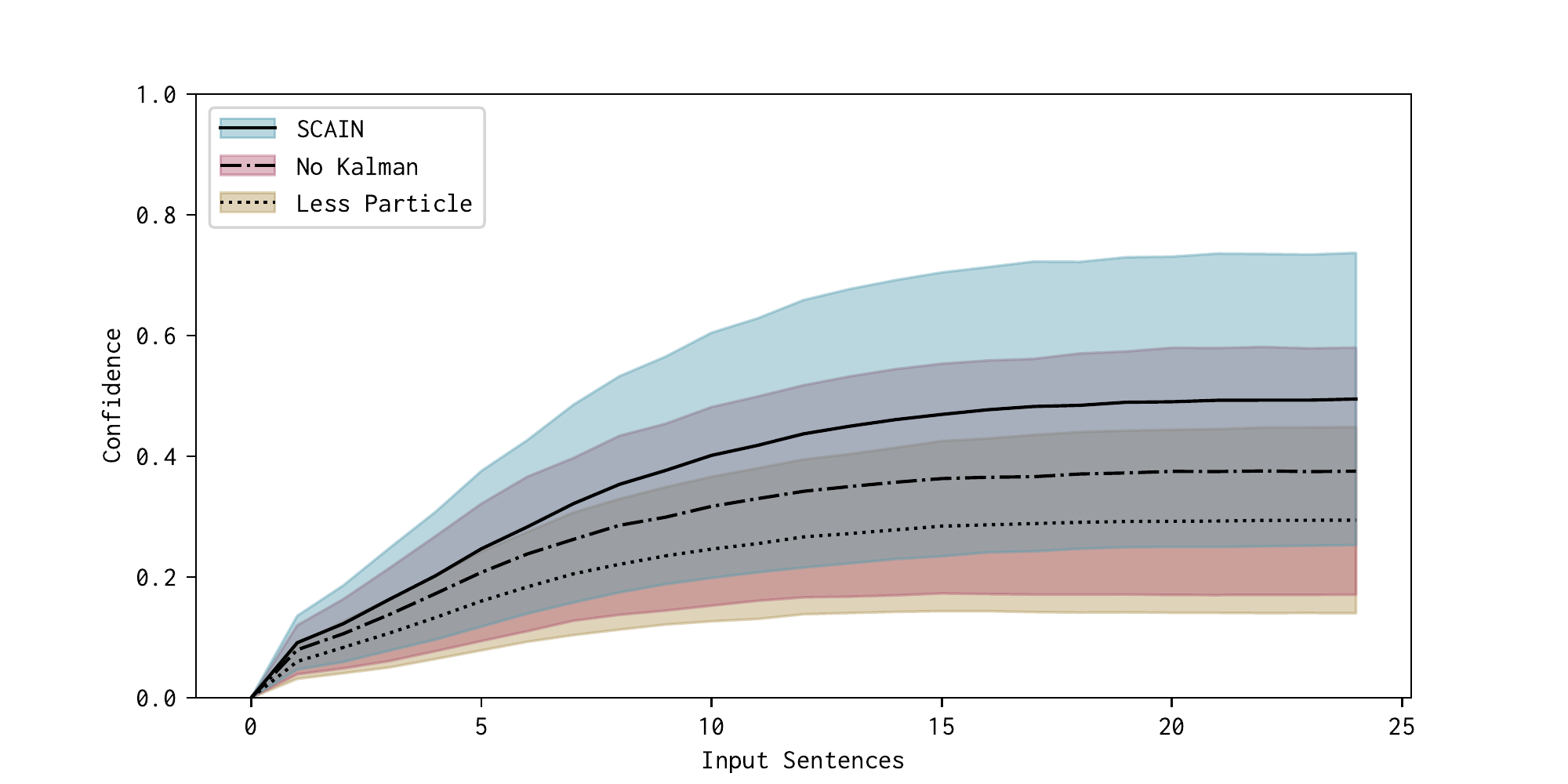}
  \caption[Expriment result of conversation dataset's new interpretaion task with other models]{Average and variance transitions of confidence with an increasing number of input sentences for the conversation dataset on the new interpretation task, including comparison with the other models.}
  \label{fig_eval_conv_cand_inner}
\end{minipage}
\end{figure}

\Subsubsection{Evaluation Result}
Figures~\ref{fig_eval_conv_text_human} and~\ref{fig_eval_conv_text_inner} show graphs of the transition of confidence in interpreting the polysemous word that the presenters used in the interpretation estimation task.
We compare the {\propmethod} and human results in Fig.~\ref{fig_eval_conv_text_human}.
Here, the human confidence was the value that the estimator input with the sliders during the conversation.
{\propmethod} is compared with the No Kalman and Fewer Particles models in Fig.~\ref{fig_eval_conv_text_inner}.
As for the confidence transitions in the new interpretation task, Fig.~\ref{fig_eval_conv_cand_human} shows the results for {\propmethod} when it did not have the candidate of the correct interpretation before the sentences were input, together with the human results.
The graph for the estimators in Fig.~\ref{fig_eval_conv_cand_human} shows the case in which they already had the correct interpretations.
Finally, Fig.~\ref{fig_eval_conv_cand_inner} shows the comparison among the {\propmethod} algorithm and the ``No Kalman'' and ``Fewer Particles'' models when they did not have the correct candidate at the beginning of the new interpretation task.

In addition, Table~\ref{table_eval_conv_accuracy} lists the accuracies at the end of the conversation.
Table~\ref{table_eval_conv_text_f1} lists the precision, recall, and F-measure of each model with respect to the human confidences at the end of the conversation for the interpretation estimation task. Table~\ref{table_eval_conv_cand_f1} lists the same results for the new interpretation task.

\begin{table}[htb]
\centering
\caption[Accuracy in the conversation dataset]{Accuracy for the conversation dataset, with the majority choice in terms of confidence taken as correct.}
\begin{tabular}{c|c|c}
  Model & Interpretation estimation task & New interpretation task \\ \hline
  {\propmethod}     & 0.722 & 0.411 \\
  No Kalman         & 0.578 & 0.378 \\
  Fewer Particle    & 0.089 & 0.244 \\
  Human (Estimator) & 0.633 & ---   \\
\end{tabular}
\label{table_eval_conv_accuracy}
\end{table}

\begin{table}[htb]
\centering
\caption[Precision, Recall, F-measure of the conversation dataset in interpretation estimation task]{Precision, recall, and F-measure for the conversation dataset with respect to human confidences in the interpretation estimation task.}
\begin{tabular}{c|ccc}
  Model             & Precision & Recall & F1    \\ \hline
  {\propmethod}     & 0.754     & 0.860  & 0.803 \\
  No Kalman         & 0.692     & 0.632  & 0.661 \\
  Fewer Particle    & 0.500     & 0.070  & 0.123 \\
\end{tabular}
\label{table_eval_conv_text_f1}
\end{table}

\begin{table}[htb]
\centering
\caption[Precision, Recall, F-measure of the conversation dataset in new interpretation task]{Precision, recall, and F-measure for the conversation dataset with respect to human confidences in the new interpretation task.}
\begin{tabular}{c|ccc}
  Model             & Precision & Recall & F1    \\ \hline
  {\propmethod}     & 0.703     & 0.456  & 0.553 \\
  No Kalman         & 0.647     & 0.386  & 0.484 \\
  Fewer Particle    & 0.727     & 0.281  & 0.405 \\
\end{tabular}
\label{table_eval_conv_cand_f1}
\end{table}

\Section{Discussion}
\Subsection{Wikipedia Dataset}
In the evaluation with the Wikipedia dataset, Figs.~\ref{fig_eval_wiki_text} and~\ref{fig_eval_wiki_cand} indicate that {\propmethod} exceeded the other models for both the interpretation estimation task and the new interpretation task at almost all times throughout the sentence inputs.
In particular, {\propmethod} showed a remarkable characteristic of the dynamic interpretation of sentences as the context progressed.
Although the No Kalman model exceeded {\propmethod} at the early stage of conversation in the interpretation estimation task, the confidence of {\propmethod} rose with the input of more sentences, while the increase in confidence of the No Kalman model was small.
The No Kalman model does not update the distributed representation and continues to use its initial knowledge.
In contrast, because {\propmethod} considers {\interdependence}, it continues to modify the distributed representations contained in the input sentences.
Therefore, we conclude that the No Kalman model exceeded {\propmethod} at the early stage because the updating of the distributed representations did not work effectively at the early stage with poor contextual information.
After the contextual information had been fully developed in the later stages, it was possible to estimate the correct interpretation through the updating.
In addition, the confidence of the Fewer Particles model was always lower than that of the other models.
The results confirmed that, even if the distributed representations are updated, the interpretation cannot be estimated correctly with an insufficient number of particles.
It is thus necessary to search in parallel with multiple particles.

In the new interpretation task, the confidence tended to rise for all models as sentences were input.
The average of the {\propmethod} algorithm's final confidence was 0.4, whereas those of the No Kalman and Fewer Particles models were 0.2.
Although it was not possible to obtain as high confidence as in the interpretation estimation task, when all the models knew all the interpretation candidates in advance, we confirmed that parallel update of the distributed representation and search with a sufficient number of particles were also effective for acquiring a new interpretation.

Regarding the accuracy, Table~\ref{table_eval_wiki} shows that {\propmethod} achieved 0.841 for the interpretation estimation task and 0.410 for the new interpretation task, which was much larger than the accuracies for the No Kalman and Fewer Particles models.
The accuracies of those models were all extremely small, at 0.2 or less, because the threshold of accuracy was set to 0.5 to achieve a majority, whereas the averages of the final confidences were below 0.4.
Likewise, the accuracy threshold diminished the {\propmethod} algorithm’s accuracy for the new interpretation task to {0.410}, which was lower than {0.841} for the interpretation estimation task.

In the Wikipedia dataset, only other's utterance $z$ was contained in the input sentences, and own utterance $u$ was not handled.
In {\propmethod}, own utterance $u$, however, affects only context $x$, and its updating equation is the same as the other's utterance's one except for the coefficient of the contribution rate.
Although own utterance $u$ helps context $x$ converge faster, we presume the above characteristics occur irrespective of whether the inputs of sentences include own utterance $u$.

%

\Subsection{Conversation Dataset}
For the interpretation estimation task with the conversation dataset, Fig.~\ref{fig_eval_conv_text_human} indicates that {\propmethod} could output the same level of interpretation confidence as humans could, although its confidence was lower than that of the humans on average.
Actually, in Table~\ref{table_eval_conv_accuracy} the {\propmethod} algorithm’s accuracy of 0.72 exceeded the human accuracy of 0.633.
We can consider several reasons why the human accuracy was not so high on average and had large variance. In some cases it was hard for the estimators to limit themselves to some specific interpretations, because the presenters did not provide much information. In other cases the estimators stuck to eliminating the correct interpretation according to their intuition. In addition, the estimators sometimes tried to infer the result logically according to the semantic relations between words, unlike {\propmethod}, which could use only the word occurrences.
Moreover, the humans could choose extreme values such as 0\% and 100\% with the slider input, which caused a large variance.
As for why the human confidence exceeded that of {\propmethod} in the second half of the conversation even though {\propmethod}'s accuracy exceeded that of the humans, we assume that the human variance increased in the latter half, while {\propmethod} could always easily take intermediate values because of its parallel search mechanism.
Likewise, although the variance of the human confidence was small in the first half of the conversation, {\propmethod} showed a tendency to draw estimation conclusions with less contextual information.
In comparison with the other models in Fig.~\ref{fig_eval_conv_text_inner}, although the variances were significant, {\propmethod} exceeded the other two models in terms of the average value.
In addition, the F-measure results for each model in Table~\ref{table_eval_conv_text_f1} indicate the advantage of {\propmethod}, which had a value of 0.803, exceeding 0.663 for the No Kalman model and 0.123 for the Fewer Particles model.
Unlike the evaluation with the Wikipedia dataset, the performance of {\propmethod} exceeded that of the No Kalman model even in the early stage, because the probabilities of word occurrences in the human conversation sentences differed from those for the Wikipedia articles that were used for learning of prior knowledge.
In human conversation, people often describe subjective impressions, whereas such impressions hardly appear in an encyclopedia such as Wikipedia.
Therefore, we presume that {\propmethod} surpassed the No Kalman model because updating the distributed representation with the Kalman filter worked effectively from the early stage of the conversation.

For the new interpretation task, it was difficult to collect the humans' acquisition process for new interpretations through conversation. Hence, Fig.~\ref{fig_eval_conv_cand_human} shows the graph of the human confidences for the interpretation estimation task in order to compare with the {\propmethod} algorithm’s confidences in the new interpretation task, even though the estimator had a correct candidate for the interpretation in advance.
In the early stage, there is a considerable difference between the confidence of the humans, who had prior knowledge, and that of {\propmethod}, which did not.
Although the average for {\propmethod} never exceeded that of the humans, however, it did reach similar averages and variance in the latter half.
In comparison with the other models in Fig.~\ref{fig_eval_conv_cand_inner}, although the variances were significant, {\propmethod} exceeded the other models at all times.
This tendency is similar to the result of the interpretation estimation task.
In addition, although the differences in the correct answer rates in Table~\ref{table_eval_conv_accuracy} and the F-measures with respect to human confidence in Table~\ref{table_eval_conv_cand_f1} were not as pronounced as for the interpretation estimation task, {\propmethod} exceeded the other two models in performance.

\Section{Future works}
For not only the Wikipedia dataset but also the conversation dataset, {\propmethod} could estimate the same interpretations as humans.
There was a tendency, however, to draw a conclusion even at the initial stage as compared with human confidence, and there was a notable feature in that extreme confidence could not be output as strongly as by the humans.
In both cases, the cause was the weight calculation method in Step 4. We think that it will be possible to obtain an output similar to human confidence by introducing a calculation formula that captures further features or by applying other learning mechanisms.
Furthermore, because {\propmethod} accounts for only the distributed representation, it does not deal with the semantic relationships between words, unlike human beings.
For further performance improvement, it will be necessary to consider not only the distributed representation space of words but also spaces taking semantic information into consideration.
In our evaluation, {\propmethod} played a role as an interpreter.
However, in order for {\propmethod} to perform as a side of providing topics, it is necessary to use more complex probabilistic models.
Specifically, {\propmethod} has to handle a situation where the context in which the other interprets is different from the one which {\propmethod} intends.
From the viewpoint of the diversity of the polysemous words, we used Japanese for the evaluation.
In principle, {\propmethod} can be applied to other languages such as English. However, verification of the effectivity is a future task.
In English, there are many examples where part of speech changes, even if the label of a word is the same. Thus it is unknown whether the current {\propmethod} can handle it.
Finally, in this research, although we tried optimizing {\interdependence} by applying the SLAM algorithm, FastSLAM itself is a classical method.
As more effective and highly scalable methods are developed, it will be conceivable to design neural networks having structures based on SLAM's Bayesian network.

\Section{Conclusion}
In this paper, we focused on optimizing {\interdependence}, which means that, in human conversation, context is decided by word interpretations, and the word interpretations are determined by the context. We thus proposed a method of simultaneous contextualization and interpreting ({\propmethod}) based on the traditional SLAM algorithm, by associating the self-position and environment map of SLAM with the context and word interpretations, respectively, in a distributed representation space.
The {\propmethod} algorithm achieves online optimization by dynamically estimating the context and interpretation of polysemous words for sequential input of sentences as in conversation, and it can obtain new interpretations of words during such conversation.
For evaluation, we created a dataset from Japanese Wikipedia's disambiguation pages and measured whether {\propmethod} could estimate word interpretations correctly under the circumstances of sequentially inputting sentences.
As a result, {\propmethod} achieved the accuracy of 84\% at the end of the sentences, and even when {\propmethod} did not have previous knowledge of the correct interpretations, new interpretations were obtained in 41\% of cases.
Furthermore, to evaluate whether {\propmethod} worked as expected for real human conversation, we created a conversation dataset with experimental collaborators.
Evaluation results for the conversation dataset showed that the accuracy of {\propmethod} exceeded that of humans, and the algorithm's capability was confirmed.

\Section*{Acknowledgments}
This work was supported by JST CREST Grant Number JPMJCR19A1, Japan.

\bibliographystyle{unsrt}
\bibliography{reference}

\end{document}